\documentclass{llncs}

\usepackage{todonotes}
\usepackage{amssymb}
\usepackage{graphicx}
\usepackage{amsmath}
\usepackage{sidecap}
\usepackage{wrapfig}
\usepackage{tabularx,booktabs}
\usepackage{enumitem}
\usepackage{color, colortbl}

\usepackage{url}
\newcommand{\keywords}[1]{\par\addvspace\baselineskip
\noindent\keywordname\enspace\ignorespaces#1}
\newcommand{\sd}[1]{{\tiny (#1)}}

\begin{document}
\sloppy

\mainmatter  

\title{Novelty Search in Competitive Coevolution}

\titlerunning{Novelty Search in Competitive Coevolution}

\author{Jorge Gomes$^{1,2}$ \and Pedro Mariano$^2$ \and Anders Lyhne Christensen$^{1,3}$}
\authorrunning{Gomes et al.}

\institute{$^1$Instituto de Telecomunica\c{c}\~{o}es, Lisbon, Portugal \\
$^2$LabMAg -- Faculdade de Ci\^{e}ncias da Universidade de Lisboa, Portugal \\
$^3$Instituto Universit\'{a}rio de Lisboa (ISCTE-IUL), Lisbon, Portugal \\
\url{jgomes@di.fc.ul.pt}, \url{plmariano@fc.ul.pt}, \url{anders.christensen@iscte.pt}
}
%
%

\maketitle

\begin{abstract}
One of the main motivations for the use of competitive coevolution systems is their ability to capitalise on arms races between competing species to evolve increasingly sophisticated solutions. Such arms races can, however, be hard to sustain, and it has been shown that the competing species often converge prematurely to certain classes of behaviours. In this paper, we investigate if and how novelty search, an evolutionary technique driven by behavioural novelty, can overcome convergence in coevolution. We propose three methods for applying novelty search to coevolutionary systems with two species: (i)~score both populations according to behavioural novelty; (ii)~score one population according to novelty, and the other according to fitness; and (iii)~score both populations with a combination of novelty and fitness. We evaluate the methods in a predator-prey pursuit task. Our results show that novelty-based approaches can evolve a significantly more diverse set of solutions, when compared to traditional fitness-based coevolution.
\end{abstract}

\section{Introduction}


In a coevolutionary system, two or more populations simultaneously evolve, and evaluations are conducted with an individual from each population. In the case of competitive coevolution systems (CCESs), the populations represent competing species, where each succeed at the expense of the opposing species. CCESs are appealing for a number of reasons~\cite{watson01,nolfi98}: (i)~they are suited to domains where a notion of absolute fitness might not exist; (ii) each population acts as a progressively more challenging opponent to the other population, and (iii) a CCES may be less prone to stagnation than non-coevolutionary methods, because of the ever-changing fitness landscape.

In practice, however, many of these advantages are frequently not observed~\cite{popovici12}. The key to successful coevolutionary learning is a continuous arms race between the opposing species~\cite{ficici98}. However, such an arms race is not easy to establish~\cite{ebner10}, and even when an arms race does occur, there is no guarantee that it will lead to good solutions. A number of techniques to help sustain an arms race have been proposed~\cite{rosin97}, but most of them are focused on maintaining performance against opponents from earlier generations. While such techniques can contribute to a continuous and objective improvement of the solutions, they do not guarantee that a wide range of solutions are discovered, which is one of the most appealing arguments for the use of competitive coevolution~\cite{nolfi98}. The goal of a CCES is not necessarily to find the (near-)optimal solution for a problem~\cite{popovici12}. The discovery of diverse solutions might be equally valuable, since there is no absolute measure of quality. Consider, for instance, the evolution of agents to play against human players: having a good diversity of playing strategies might be more valuable than having a single near-optimal player~\cite{yannakakis07}. Moreover, it has been shown that promoting diversity in a population can potentially lead to improvements in the population's generalisation capacity~\cite{chong09}.

Evolutionary techniques based on behavioural diversity, such as novelty search (NS)~\cite{lehman11}, have shown effective in overcoming behavioural convergence. These techniques drive evolution towards behavioural innovation, often resulting in a more effective evolutionary process that produces a greater diversity of solutions compared with traditional fitness-based evolution. Although most studies focus on non-coevolutionary domains~\cite{mouret12,gomes13}, we recently showed that NS can be successfully used to overcome stable states in cooperative coevolution~\cite{gomes14}.

In this paper, we study the adaptation of behavioural diversity techniques to competitive coevolution. This combination of techniques is especially challenging because in a CCES, all populations are required to be fairly competitive at all times, which may conflict with the diversity objective~\cite{ashlock04}. We propose three different ways to apply novelty search to a CCES, and compare them with traditional fitness-based coevolution in a predator-prey pursuit task. We assess the proposed techniques along three dimensions: (i)~the quality of the best solutions achieved; (ii)~the exploration of the behaviour space; and (iii)~the diversity of high-quality solutions evolved.

\section{Related Work}

\subsection{Premature Convergence in Competitive Coevolution}


Many reports of convergence of coevolving populations to undesirable regions of the solution space can be found in the literature (for examples, see~\cite{dziuk11,avery10,reisinger07}). One of the main causes for failure is that the coevolutionary process easily gets trapped in a \emph{mediocre stable state}~\cite{ficici98}: a cycle where a limited set of solutions is adopted by the populations over and over again. This cycle can give the impression of competition, without actually causing the evolutionary process to explore new solutions or to improve the quality of the individuals.


Another reason for the loss of diversity in the populations can be the lack of a fitness gradient~\cite{watson01}. It is possible that one population becomes so dominant that it turns into an \emph{unhittable target} for its competitors. As such, the selection pressure disappears, and the populations stop improving. Ashlock et al.~\cite{ashlock04} tried to overcome the lack of a fitness gradient by rewarding individuals that generate a high variability of fitness scores in the opposing population. The approach was, however, unsuccessful, since individuals that allow for a high variability of fitness typically do not pose a significant challenge.

A number of strategies for cultivating fruitful arms races have been proposed. Rosin \& Belew~\cite{rosin97} propose three techniques that aim to select a diverse and challenging set of competitors to test the individuals: (i)~competitive fitness sharing, (ii)~shared sampling, and (iii)~the hall of fame. Other works have focused on the characteristics of the task setup that are favourable to the emergence and sustainability of arms races~\cite{nolfi12}. It has also been shown that shaping the environment or the fitness function throughout evolution can help avoid convergence to uninteresting and non-diverse solutions~\cite{dziuk11}.

\subsection{Novelty Search}

Novelty search (NS)~\cite{lehman11} is an evolutionary technique in which the population is driven towards behavioural novelty. NS has the potential to avoid premature convergence, and evolve a wide diversity of solutions in a single evolutionary run, as opposed to fitness-driven evolution that typically converges to a certain class of solutions~\cite{gomes13}. The approach has been applied in the domain of evolutionary robotics and evolution of agent controllers with considerable success~\cite{mouret12,gomes13,gomes14}.


\vspace{-10pt}
\subsubsection{Novelty search algorithm}
\label{sec:ns}
The distinctive aspect of NS is how the individuals of the population are scored. Instead of scoring individuals according to how well they perform a given task, which is typically measured by a fitness function, individuals are scored based on their behavioural novelty. Behavioural novelty is measured with a \emph{novelty metric} that quantifies how different an individual is from other, previously evaluated individuals. The distance between two individuals is typically given by the distance between their \emph{behaviour characterisation vectors}. These vectors are commonly composed of behavioural traits that the experimenter considers relevant for the task.

To measure how far an individual is from other individuals in behaviour space, the novelty metric relies on the average behaviour distance of that individual to the $k$ nearest neighbours. Potential neighbours include the other individuals of the current generation and a random sample of individuals from previous generations (stored in an archive). Candidates from sparse regions of the behaviour space therefore receive higher novelty scores, which results in a constant evolutionary pressure towards behavioural innovation. 

\vspace{-10pt}
\subsubsection{Combining novelty and fitness}

As NS is guided by behavioural innovation alone, its performance can be greatly affected by the size and shape of the behaviour space~\cite{gomes13}. Since NS is essentially an exploratory technique, most of the effort may be spent in behaviour regions that are irrelevant for the fulfilment of the task. Therefore, NS is often combined with fitness-based evolution to promote exploration of high-fitness behaviour regions.

In our experiments, we use a variant of \emph{progressive minimal criteria novelty search} (PMCNS)~\cite{gomes12} to combine novelty and fitness. PMCNS restricts exploration to behaviour regions associated with relatively high fitness scores. At each generation $g$, the individuals' selection scores are assigned in the following way: if the fitness score of an individual is above the minimal criterion $mc_g$, its novelty score alone is used for selection, otherwise it receives a score of zero. The $mc_g$ criterion corresponds to the $P$-th percentile value of the fitness scores of all individuals of generation $g$.

\section{Approach}
The application of behavioural diversity techniques to a CCES poses a number of challenges. In order to evolve effective solutions, coevolution requires an arms race between the coevolving species. This implies that all species are simultaneously improving to defeat one another: they learn to exploit flaws in the opponents' strategy. Behavioural diversity techniques are, however, essentially exploratory: evolution is driven towards behavioural novelty, not necessarily better solutions (i.e, solutions that are able to defeat the opponents). Prioritising diversity instead of competitiveness can compromise the effectiveness of coevolution, as shown in~\cite{ashlock04}. On the other hand, it has also been shown that when a population has converged to a strategy, it can be beneficial to reward alternate strategies, in order to foster the evolution of new behavioural traits~\cite{dziuk11}.


We study the application of NS to a CCES with two competing species. The individuals' fitness score 
is measured in the traditional way: by competing against a representative sample of the individuals from the opposing species. In these competitions, we record the individual's behaviour as a task-specific behaviour characterisation vector. The novelty score  of an individual is then calculated by computing the distance between its characterisation vector and the vectors of current and past individuals from the same species. The following methods are evaluated in this paper:
\setdescription{leftmargin=1.77cm,labelindent=0pt}\begin{description} [labelwidth=1.6cm]
\item[Fit] Selection based exclusively on the fitness score, in both populations. This method is used as a baseline.
\item[NS-Both] Selection based exclusively on the novelty score in both populations. 
\item[NS-\emph{p}] Selection based on the novelty score in population $p$ (in our experiments, either \emph{Pred} for predator, or \emph{Prey}), and based on the fitness score in the other population.
\item[PMCNS] Both populations are scored with PMCNS based on both novelty and fitness scores. The individuals are rewarded for displaying behavioural novelty while still meeting the minimal fitness criterion.
\end{description}

\section{Experimental Setup}

\subsection{Predator-Prey Pursuit}

\begin{wrapfigure}{l}{0.35\textwidth}
\vspace{-20pt}
\centering
\includegraphics[width=0.35\textwidth]{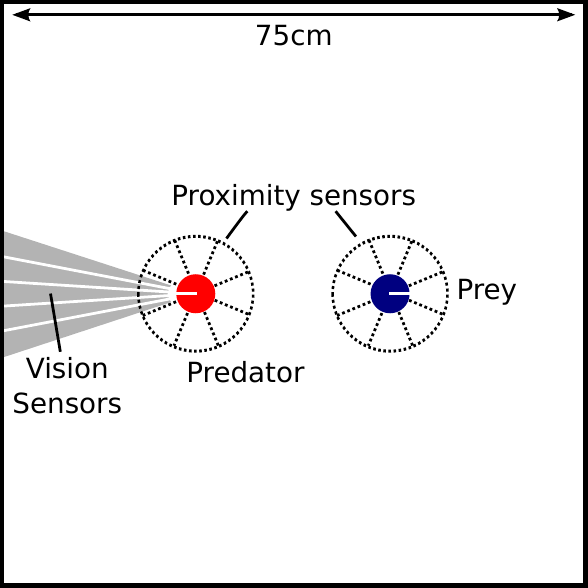}
\vspace{-15pt}
\caption{Setup of the predator-prey task. The agents are initially placed in fixed positions, facing opposing directions.}
\vspace{-10pt}
\label{fig:setup}
\end{wrapfigure}

The predator-prey pursuit task is a common testbed for CCESs~\cite{nolfi12,cliff95}. In this task, two agent controllers are coevolved: one for a predator and one for a prey. The predator's objective is to capture (touch) the prey, and the prey's objective is to escape the predator. The agents operate in a closed square arena of 75\,x\,75\,cm$^2$. A simulation trial ends if the predator captures the prey or if 100 seconds elapse. The initial conditions and setup of the task are depicted in Figure~\ref{fig:setup}.

The experimental setup is based on~\cite{nolfi98}. The prey and the predator move at a maximum speed of 5\,cm/s. Both the prey and predator have eight proximity sensors, evenly distributed on their bodies, with a maximum range of 5\,cm. The proximity sensors can detect both the walls and the other agent. The predator is additionally equipped with five vision sensors to detect the prey. The vision sensors have unlimited range and when combined provide a view angle of 40$ ^{\circ}$. All sensors are binary: they return 1 if something is detected, and 0 otherwise. Each agent is controlled by a feed-forward neural network. Sensor readings are fed to the network, and two outputs control the speed of the two wheels. The networks of the predator and the prey have respectively 7 and 5 hidden neurons. 

The fitness of the prey is given by the fraction of simulations it avoided capture. The fitness of the predator is the opposite: the fraction of simulations it was able to capture the prey. The behaviour characterisation used in the novelty-based setups is the same for both the predator and the prey. It captures behavioural traits that are intuitively relevant in the context of the task. The characterisation is a vector of four elements: (i)~the simulation length; (ii)~the mean distance to the other agent throughout the simulation;~(iii) the mean agent movement speed; and~(iv) the mean distance between the agent and the closest wall. All elements are normalised to $[0,1]$.

\subsection{Coevolutionary Algorithm}

The predator and prey controllers are coevolved in two separate populations. The weights of the neural controllers are directly encoded in the chromosomes. Each population is evolved with a simple evolutionary strategy with the following parameters: no crossover, a gene mutation rate of 5\%, population size of 200, and tournament selection of size 5. At the end of each generation, the individual with the highest fitness score of each population is added to the respective \emph{hall of fame}~\cite{rosin97}. The individuals of each population are evaluated against a set of ten competitors randomly drawn from the opposing species' hall of fame. 

NS is implemented as described in Section~\ref{sec:ns}. Each population has its own novelty archive. Individuals are added to the archive with a probability of 3\%. The archive can hold at most 1000 individuals: after reaching the limit, random individuals are removed to allow space for new ones. PMCNS was configured with a percentile value $P$ of 0.5, i.e., the median fitness of the population. Values of 0.25 and 0.75 were also tested, but yielded inferior performance.  

\section{Results}

\subsection{Quality of the Solutions}

\begin{figure}[b!]
\centering
\includegraphics[width=1\linewidth]{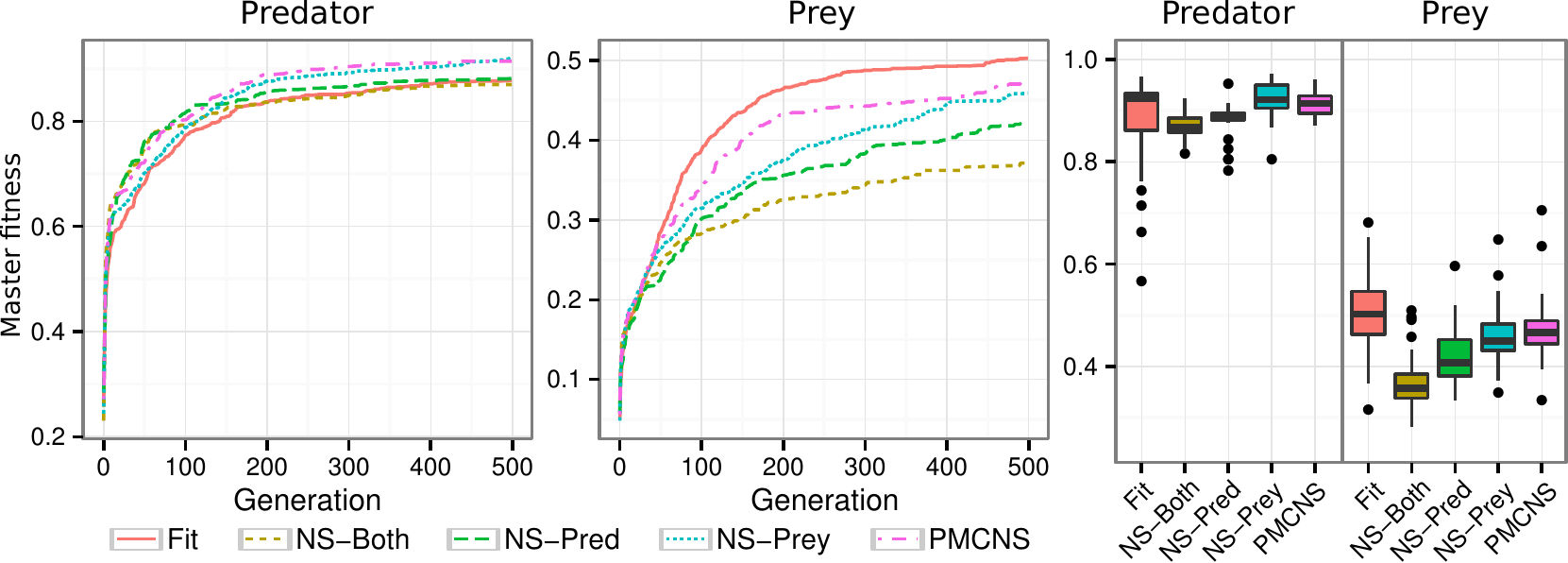}
\caption{Left and middle: best fitness found so far at each generation, for each population and each method. The fitness values correspond to the \emph{master fitness}, and are averaged over 30 evolutionary runs. Right: boxplots of the highest fitness scores found in each evolutionary run.}
\label{fig:performance}
\end{figure}

In all evolutionary runs, the individuals with the highest fitness score from both populations in every generation were recorded. To obtain an objective measure of the quality of these individuals, we performed a two-step master tournament~\cite{nolfi98}, obtaining a \emph{master fitness} and a \emph{master behaviour characterisation}. First, the best individuals of each evolutionary run were identified by evaluating them against the individuals of the opposing species from the same run. Second, the individuals were evaluated against all the individuals identified in the first step, from all evolutionary treatments. The analysis of the individuals' master fitness can be seen in Figure~\ref{fig:performance} and Table~\ref{tab:summary} (Best fitness column).

These results show that all evolutionary treatments were able to achieve high-quality predator controllers. The differences between the treatments are relatively small. \emph{Fit} is only significantly different from \emph{NS-Both}, which is significantly inferior to all other treatments (Mann-Whitney U test, $p$-value $< 0.01$). The evolution of prey controllers, on the other hand, revealed larger differences between the evolutionary treatments. The treatment in which novelty had the biggest influence (\emph{NS-Both}) was the lowest performing treatment with respect to the quality of the solutions ($p$-value $< 0.01$). Conversely, the treatments with highest fitness pressure yielded the highest scoring solutions, with none of the novelty-based treatments outperforming \emph{Fit} ($p$-value $< 0.01$). This result confirms that in order to succeed, the populations need to have a strong selection pressure towards defeating the opposing population. In the predator-prey task, using NS to promote exploration of the behaviour space did not yield significant advantages regarding the achieved fitness scores.

\subsection{Behaviour Space Exploration}

One of the main advantages of novelty-based evolutionary techniques is the diversity of evolved solutions. To make a quantitative analysis of the behaviour space exploration, the space was first divided in regions of equal size: each dimension of the behaviour characterisation was discretised into 5 levels, resulting in a total of 625 regions. We then calculated how many times each region was visited throughout each evolutionary run (based on the \emph{master characterisations}). This distribution was compared with the uniform distribution to obtain a measure of behaviour space coverage. The non-visited regions of the behaviour space (not reached by any method) were excluded, see Table~\ref{tab:summary} -- Global exploration. To measure the exploration of the high-fitness behaviour regions, we followed the same procedure but excluded the regions where no reasonably good solutions were found (master fitness below 0.8 for the predator population, and 0.3 for the prey population), see Table~\ref{tab:summary} -- Elite exploration.

\definecolor{lightgray}{RGB}{235, 235, 235}
\newcommand{\shaderow}{\rowcolor{lightgray}[2pt][2pt]}

\begin{table}[tb]
\caption{Best fitness found in each evolutionary run; Average coverage of the behaviour space in a whole evolutionary run; Average coverage of the high-fitness behaviour regions. Values are averages of 30 evolutionary runs for each method. Standard deviations are in parentheses.}
\begin{center}
\begin{tabularx}{1\textwidth}{X X X X X X X}
\toprule
 & \multicolumn{2}{c}{Best fitness} & \multicolumn{2}{c}{Global exploration} & \multicolumn{2}{c}{Elite exploration} \\
 \cmidrule{2-7}
 & \multicolumn{1}{c}{Predator} & \multicolumn{1}{c}{Prey} & \multicolumn{1}{c}{Predator} & \multicolumn{1}{c}{Prey} & \multicolumn{1}{c}{Predator} & \multicolumn{1}{c}{Prey} \\
 \midrule
\shaderow \emph{Fit} & 0.88 \sd{0.10} & 0.50 \sd{0.09} & 0.42 \sd{0.05} & 0.37 \sd{0.05} & 0.49 \sd{0.11} & 0.46 \sd{0.06} \\ 
\emph{NS-Both} & 0.87 \sd{0.03} & 0.37 \sd{0.06} & 0.64 \sd{0.03} & 0.60 \sd{0.03} & 0.62 \sd{0.04} & 0.60 \sd{0.05} \\ 
\shaderow \emph{NS-Pred} & 0.88 \sd{0.04} & 0.42 \sd{0.06} & 0.64 \sd{0.02} & 0.40 \sd{0.04} & 0.67 \sd{0.06} & 0.43 \sd{0.08} \\ 
\emph{NS-Pre}y & 0.92 \sd{0.04} & 0.46 \sd{0.06} & 0.43 \sd{0.03} & 0.59 \sd{0.03} & 0.50 \sd{0.04} & 0.70 \sd{0.04} \\ 
\shaderow \emph{PMCNS} & 0.91 \sd{0.03} & 0.48 \sd{0.07} & 0.59 \sd{0.03} & 0.53 \sd{0.04} & 0.70 \sd{0.03} & 0.59 \sd{0.06} \\
\bottomrule
\end{tabularx}
\end{center}
\label{tab:summary}
\vspace{-20pt}
\end{table}

\emph{Fit} has the lowest degree of exploration, which suggests that it typically converges to a single class of solutions. \emph{NS-Both}, on the other hand, has the highest degree of global exploration, but the same does not hold for elite exploration. The lack of elite exploration can explain the inferior performance of \emph{NS-Both} regarding the achieved fitness scores: too much effort is spent on the exploration of low-fitness behaviour regions. The results show that \emph{PMCNS} is able to avoid spending too much effort on exploration of low-fitness behaviours. The global exploration in PMCNS is still significantly higher than \emph{Fit}, and it has one of the highest degrees of elite exploration.

Comparing \emph{NS-p} to \emph{Fit}, it is possible to observe that the fitness-driven population in \emph{NS-p} has similar exploration and fitness scores (see in Table~\ref{tab:summary} -- Prey: \emph{NS-Pred}, \emph{Fit}, and Predator: \emph{NS-Prey}, \emph{Fit}). The fitness-driven population is not significantly affected by the greater behavioural diversity in the opposing, novelty-driven population. 
On the other hand, if we compare the novelty-driven population with \emph{NS-Both}, we can see that it is favourable to have one of the populations driven by fitness {(see in Table~\ref{tab:summary} -- Predator: \emph{NS-Pred}, \emph{NS-Both}, and Prey: \emph{NS-Prey}, \emph{NS-Both})}. When a novelty-driven population competes against a fitness-driven population, it is able to achieve significantly higher elite exploration scores, which also translates into higher fitness scores reached. By coevolving with a highly competitive, fitness-driven population, the novelty-driven population tends to move towards better solutions -- even though their individuals are selected based on their novelty scores alone.

\subsection{Diversity of Effective Solutions}

\begin{figure}[b!]
\vspace{-5pt}
\centering
\includegraphics[width=1\linewidth]{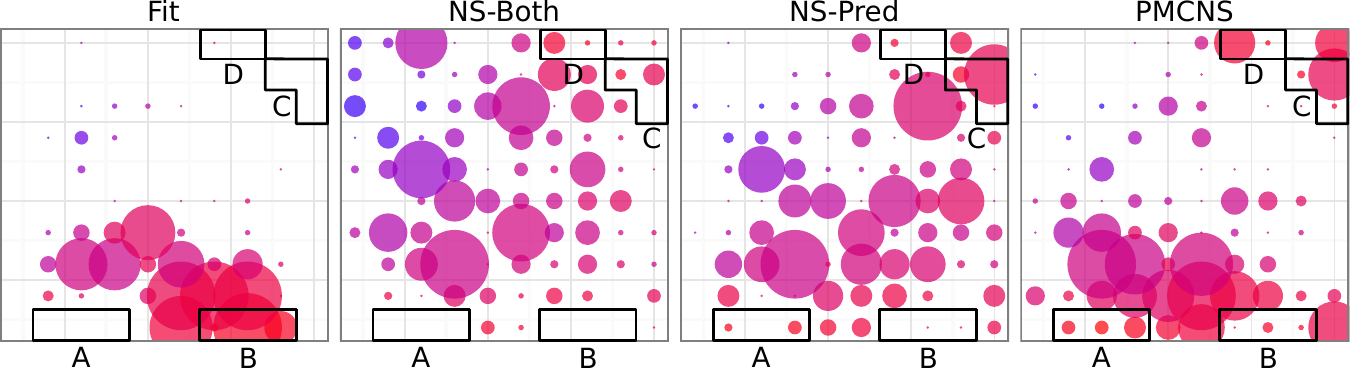}
\caption{Exploration of the predator behaviour space in typical evolutionary runs. The diameter of each circle is proportional to the number of individuals found belonging to that behaviour region. The highlighted regions A, B, C and D are associated with the highest fitness scores.}
\label{fig:exploration}
\end{figure}

The behaviour of the most effective preys is always very similar: they move at full speed in circular trajectories around the arena, using the proximity sensors to avoid the walls and to escape the predators when being chased. This behaviour typically fails when the predator quickly approaches the prey head on or from the sides, or when the prey is being chased and gets trapped between a wall and the predator. The set of successful predator strategies evolved was more diverse, especially in novelty-based treatments. To visualise the diversity of predator behaviours, we reduced the four dimensions of the characterisation using a Kohonen self-organising map. The Kohonen map was trained using a sample of the predator behaviours (\emph{master characterisations}) found in all evolutionary runs. The solutions evolved in each run were then mapped individually. Figure~\ref{fig:exploration} shows one typical evolutionary run of each treatment. \emph{NS-Prey} is omitted since the exploration in the predator population is similar to \emph{Fit}.



The results showed by the exploration maps are in accordance with the exploration scores (Table~\ref{tab:summary}). The evolutionary runs of \emph{Fit} typically converge to region B (23 out of 30 runs). In seven of those runs, solutions in region A were also discovered. Only one run of \emph{Fit} explored regions C and D, although these regions are associated with solutions of similar quality to the solutions of regions A and B. \emph{NS-Both} explored the behaviour space relatively uniformly, without any noticeable bias towards specific regions. \emph{NS-Pred} is similar to \emph{NS-Both}, but clearly spends more effort in high-fitness regions. The exploration pattern of the \emph{PMCNS} treatment is in between that of \emph{NS-Both} and \emph{Fit}: a single evolutionary run typically converges to multiple classes of high-quality solutions.

Observing the predator solutions in action confirms that the novelty-based treatments can evolve an interesting and diverse set of behaviours. The following behaviour descriptions correspond to the highest scoring predator solutions found in each of the behaviour regions highlighted in Figure~\ref{fig:exploration}.
\begin{enumerate}[label=(\Alph*)]
\item Moves backwards and in a curved trajectory until the prey is detected with the vision sensors. Afterwards, the predator stops and rotates to face the prey. When the prey gets close, the predator rushes towards it.
\item Moves forwards at full speed, avoiding the walls, until the vision sensors detect the prey. Afterwards, the predator chases the prey at full speed.
\item Moves slowly and directly towards a wall, and then stays there, facing the centre of the arena. When a prey passes nearby, the predator quickly rushes forward to catch the prey.
\item Does not initially move, only rotates to find the prey. When the prey is located, the predator keeps rotating to face the prey, and moves backwards very slowly. When the prey is near, the predator rushes towards it.
\end{enumerate}

\section{Conclusions}

We proposed three methods to promote behavioural diversity in competitive coevolution: selection based on novelty score in both populations (\emph{NS-Both}), novelty score only in one of the populations (\emph{NS-p}), and a combination of novelty and fitness scores in both populations (\emph{PMCNS}). These methods were compared with traditional fitness-based coevolution (\emph{Fit}) in a predator-prey task. 

With respect to the highest fitness scores achieved in the evolutionary runs, there was no significant advantage of novelty-based methods, when compared to fitness-based evolution. The novelty-based methods did, however, display significantly higher degrees of behaviour space exploration. The \emph{NS-Both} method had the lowest performance among the novelty-based methods, since most of the exploration effort was spent in behaviour regions associated with relatively low fitness scores. However, when a novelty-driven population (\emph{NS-p}) competed against a fitness-driven population, results showed that the novelty-driven population had a significantly stronger tendency to explore high-fitness behaviour regions. \emph{PMCNS} was effective all-around: it could achieve fitness scores comparable to \emph{Fit}, and could discover a wide range of high-quality solutions in a single evolutionary run, for both predator and prey controllers.

The novelty-based methods could consistently explore new behaviour regions and find classes of solutions that \emph{Fit} rarely or never reached. In particular, a diverse set of interesting high-quality predator solutions was identified. Our experiments showed that novelty-based techniques can be used to avoid behavioural convergence and discover a broad diversity of solutions in competitive coevolution systems. The effectiveness of the novelty-based techniques depends on the balance between diversity and competitiveness, since the populations require challenging opponents in order to achieve solutions of high objective quality.


\subsubsection{Acknowledgements}
{\small
This research is supported by Funda\c{c}\~{a}o para a Ci\^{e}ncia e Tecnologia (FCT) grants PEst-OE/EEI/LA0008/2013, PEst-OE/EEI/UI0434/2014, SFRH/BD/89095/2012 and EXPL/EEI-AUT/0329/2013.
}

\bibliographystyle{splncs03}
\bibliography{ppsn}

\end{document}